# A design of human-like robust AI machines in object identification

**Bao-Gang Hu**[1,2,*] **and Wei-Ming Dong**[1, 2]

[1]NLPR, Institute of Automation, Chinese Academy of Science, 100091, China
[2]University of Chinese Academy of Sciences, Beijing, 100091, China
[*]hubg@nlpr.ia.ac.cn

## ABSTRACT

This is a perspective paper inspired from the study of Turing Test[1] proposed by A.M. Turing (23 June 1912 - 7 June 1954) in 1950. Following one important implication of Turing Test for enabling a machine with a human-like behavior or performance, we define human-like robustness (HLR) for AI machines. The objective of the new definition aims to enforce AI machines with HLR, including to evaluate them in terms of HLR. A specific task is discussed only on object identification, because it is the most common task for every person in daily life. Similar to the perspective, or design, position by Turing, we provide a solution of how to achieve HLR AI machines without constructing them and conducting real experiments. The solution should consists of three important features in the machines. The first feature of HLR machines is to utilize common sense from humans for realizing a causal inference. The second feature is to make a decision from a semantic space for having interpretations to the decision. The third feature is to include a "human-in-the-loop" setting for advancing HLR machines. We show an "identification game" using proposed design of HLR machines. The present paper shows an attempt to learn and explore further from Turing Test towards the design of human-like AI machines.

## Introduction

In recent years, the tremendous success of deep learning (DL) has advanced artificial intelligence (AI) for wider applications[2,3]. However, some weaknesses are appeared in applications, such as black box lacking interpretations about the outcomes of DL or AI machines[4,5]. One typical example showing the weaknesses is about identifying a panda animal (Figure 1)[6]. In an image with a panda, a DL machine was able to identify it correctly with 57.7% confidence. However, after adding a certain degree of noise, the machine identified the noised image with a gibbon animal having 99.3% confidence, and failed to give interpretations for the new decision. If asking any person to see the noised image, one can still tell the animal correctly. This example shows that the existing AI machines do not take human-like knowledge in their identifications. For overcoming this difficulty, we consider to apply the principle behind Turing Test in the design of AI machines. Turing Test was proposed based on the question "*Can machine think*", and then it was transformed into an evaluation problem for a human evaluator to judge an identity from an "*imitation game*". Hence, Turing Test is actually a design for evaluating a machine in terms of human-like intelligence. Inspired by this idea, we will go further to propose a design for enforcing a machine with human-like robustness which is defined in the next section.

## Definitions of robustness and two goals

In this paper we take most terms directly in a wider sense. For example, we consider machine to be computer, system, agent, model, etc. Only some core terms are defined, such as.

Definition 1: *Robustness* is a desirable property of a living or a machine in action correctly under any type of disturbances.

Definition 2: *Human robustness* (**HR**) refers to a class of robustness gained by humans.

Definition 3: *Machine robustness* (**MR**) refers to a class of robustness gained by machines.

Definition 4: *Human-like robustness* (**HLR**) is a class of MR shared in a union with HR.

Figure 2 shows the relations among the three specific sets (or classes) in robustness. Within a set of robustness, there still exist more subsets, such as for other living things[7]. When humans shows HR in telling a panda from the noised image in Figure 1, a machine can demonstrate its MR in watermark detections[8] from a copyright image. Note that all robustness sets are dynamic in evolutions, and HLR is defined in relation to the two goals of AI machines stated by Hu and Qu[9]:

"*Engineering goal: To create and use intelligent tools in helping humans maximumly for good.*

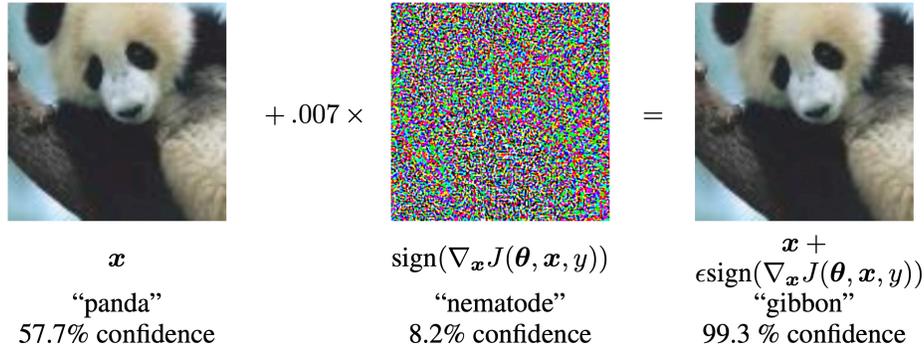

**Figure 1.** A demonstration of the "brittleness" of DL machines in object identification[6]. Left: Original image identified as panda. Middle: The noise added on the original image. Right: Noised image identified as gibbon.

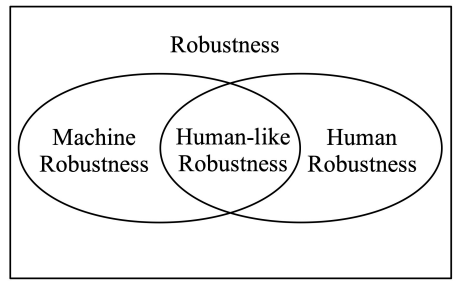

**Figure 2.** An Euler diagram of three specific sets in robustness.

*Scientific goal: To gain knowledge, better in depth, about humans themselves and other worlds.* "

The proposal of HLR shares the same goals above. When the future machines should evolve to include HR as much as possible in their MR, or to increase their HLR, we also emphasize its evolution on "*helping humans maximumly for good*", particularly for humanity[10,11].

## Three critical features of HLR machines

This paper follows the similar approach by Turing on proposing an "*initial game*"[1], which was a design of fulfilling an evaluation task in AI studies. We will propose a design for human-like robust AI machines, or HLR machines. Three critical features are suggested as high-level guidelines in the design of HLR machines below.

Feature 1: To utilize common sense from humans for realizing a causal inference.

Feature 2: To make a decision from semantic similarity for having interpretations to the decision.

Feature 3: To include a "*human-in-the-loop*" setting for advancing HLR machines.

For clarifying some core terms used in the features above, we provide their definitions below.

Definition 5: *Common sense* is a class of empirical knowledge for humans used in our daily life.

Definition 6: *Causal inference* is a class of inference that establishes cause-and-effect relationships.

Definition 7: *Semantic similarity* is a likeness metric between two semantic objects.

Definition 8: *Human-in-the-loop* (**HITL**) is a setting in a loop at which a human operator is able to insert *a priori* into an AI machine and to determine the output of the machine.

For justifying the features in HLR machines above, we take Figure 1 as an example for discussions. It is known that humans apply common sense in their daily life reasoning[12–14]. For example, after seeing Figure 1, one may identify the object first and then explain it by the following saying.

Saying 1: "*I saw two eyes in the image, the form of the eyes is the mostly distinguished characteristics for a panda, then I identify it to be a panda*".

Although the common sense may be varied with different persons, humans do utilize them in their decision making and process them mostly from a causal inference[15–17]. In Saying 1, a specific cause-and-effect relationship is based on a set of key words "*eyes, distinguished characteristics, panda*". Unfortunately, most of the existing DL machines miss the utilization of common sense and a mechanism for a causal inference in such object identifications.

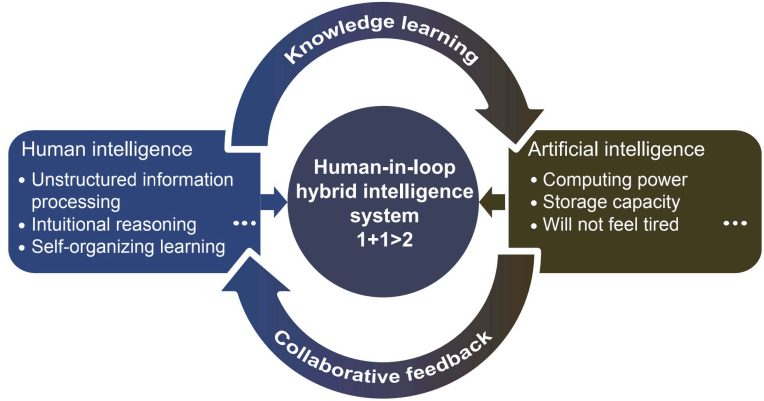

**Figure 3.** Schematic interpretation about hybrid-augmented intelligence[21].

For Feature 2, we stress that the explicit knowledge in human brain is mostly represented by a form of semantic representation[18]. Generally, semantic representations are considered to be a form of high-level knowledge[19], such as common sense. Saying 1 implies that humans apply semantic similarity in decision and provide interpretations about the decision. Hence, Feature 2 guarantees high-level knowledge with more semantic meanings in the outcome of HLR machines, rather than the only label information gained from the conventional AI machines.

For Feature 3, HITL is necessary for training in HLR machines, but may be not for testing or using. A human operator can be either a modeler or a user of machine, even for a group of operators. Two basic tasks are specified for the operator, that is, to provide instruction (including *a priori* knowledge) into the machine and to decide the final output from the AI machine. We use the term setting to reflect a physical place for humans located within such feedback loop.

## A schematic design of HLR models

The main idea of HLR machines is also inspired by a so called "*hybrid-augmented intelligence* (HAI)" proposed by Pan[20] and Zheng et al[21] respectively, for integrating both human intelligence (HI) and AI. Figure 3 illustrates an important working format of advancing intelligent level of HAI machines. We present the following mathematic description about HAI in a dynamic form:

$$HAI_{t+1} = HI_t \cup AI_t \geq HAI_t, \qquad (1)$$

where the subscript *t* represents the set at time $t(= 0,1,2,...)$. If viewing them in terms of robustness sets, we can observe that the union of HR and MR will be enlarged by their integrations in the iterations of knowledge updating. For realizing HLR to be enlarged at the same time, we propose a schematic design of HLR models in Figure 4. The term schematic indicates that the design is still at a very primary level. One can see that HLR models include both machine and human operator. We present several remarks below for explanations about the design in the context of object identification shown in Figure 1.

Remark 1: An operator can use common sense, like Saying 1, as *a priori* to insert AI machine, so that Feature 1 will be satisfied.

Remark 2: AI machine will be able to transform *a priori* knowledge from natural language into a structured representation, generate novel data, extract *a posteriori* knowledge, and make a decision from semantic similarity, so that Features 1 and 2 will be satisfied.

Remark 3: An AI machine output (MO) is a reference for humans which will include both data and extracted knowledge (or *a posteriori* knowledge).

Remark 4: After receiving MO, an human operator will examine or modify it to make a final decision as a final output (FO), so that Features 3 will be satisfied. We stress a final decision from humans for the reason that the goal "*for good*" will be technically possible for HLR models.

Remark 5: HLR models are a special class of HAI machines following an iteration procedure in Equation 1. Therefore, HLR will be increased by adding more knowledge, both implicit and explicit, via iterations.

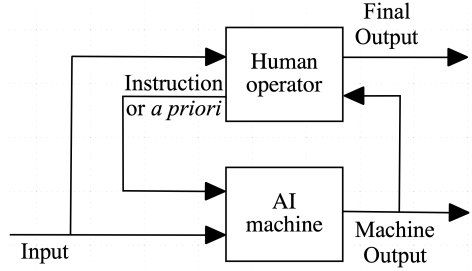

**Figure 4.** Schematic design of HLR models including an AI machine and a human operator. A machine output (MO) is set as a reference and a final output (FO) as a decision of HLR models. Both outputs include data and extracted knowledge (or *a posteriori* knowledge), but may be different. A setting of "human-in-the-loop" will ensure a human operator to control and adjust AI machine via instruction (or *a priori*) feedback.

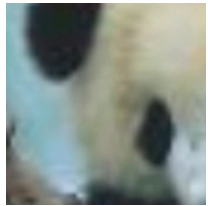

**(a)** Example 1

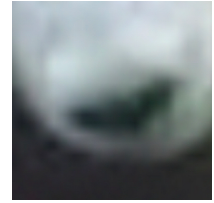

**(b)** Example 2

**Figure 5.** Two image examples in an "identification game". Human operator can guess two given images (cropped from the right image[6] in Figure 1).

## An "identification game" in using HLR models

Similar to the "*imitation game*" in Turing Test[1] which was related to some imagined dialogues between the two of three players, we also show an "*identification game*" in using HLR models. In the proposed "*identification game*", we suppose AI machine in HLR models to be an advanced machine (M), satisfying Features 1 and 2 for dealing with any type of knowledge, such as abstracting or extracting knowledge. M is capable of fulfilling many tasks based on the instruction of an human operator (H). We further suppose that, from the request of H, input data (D) is sufficiently supplied to M and M is able to preprocess D or generate novel D for M's using. For example, H can instruct M to make an image example shown in Figure 5 (a), and can provide the following saying from his/her judgment.

Saying 2: "*I still can identify it to be a panda because I can see one ear and one eye of the panda.*"

H will provide such *a priori* for M to make a training with more data. If for the given example M is able to output a correct answer satisfying Feature 2, we can say that M carries HLR in a certain degree.

The game can keep going like this, such as on Example 2 in Figure 5. At this stage, H will provide the following saying:

Saying 3: "*I cannot tell what it is about if I do not know the original image.*"

We still can test M for seeing its performance. If M can provide a correct answer as well as a meaningful interpretation, we can say that M will have a higher level of HLR over that of the operator in such object identification. The examples show that HLR models are trained at high-level, yet explicit, knowledge on both input and output. The input can be natural language, graph, or sound data, following the similar position of Turing. Note that, after the proposal of "*imitation game*", Turing commented that: "*the question and answer method seems to be suitable for introducing almost any one of the fields of human endeavour that we wish to include.*" To support this position further, we use two more examples from cartoon graphs in Figure 6. One may call them "happy cat" and "sad dog", respectively. Human identify them not only in terms of animal kinds, but also of their emotional aspects. In using MLR machines, a human operator can input common sense in training the machines for achieving such human identifications in a wider sense. After a training stage, H can still decide an HLR machine working in a form of:

$$\text{FO} = \text{MO}, \quad \text{and} \quad (A\ priori)_{t+1} = (A\ posteriori)_t, \tag{2}$$

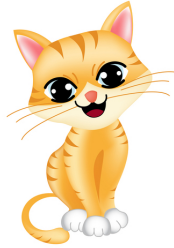
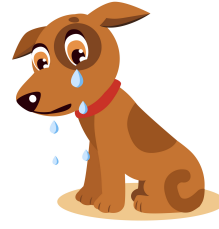

**(a)** Example 3     **(b)** Example 4

**Figure 6.** Two cartoon examples in an "identification game". One may call them "happy cat" and "sad dog", respectively. Human operator can input common sense in training MLR machines for achieving such human identifications.

in which the HLR machine will fulfill tasks without any human interaction for real-time applications, such as for controlling a fast process. In some situations, the feedback loop is removed in testing and one will get only a static HLR model without knowledge updating.

One may argue that HLR is an operator dependent. If an incorrect piece of common sense is used, its HLR models may not show the increasing evolution described by Equation 1. In this study, we suggest readers consider Equation 1 in a statistical sense over human populations. Hence, a correctness of Equation 1 relies on a condition that human knowledge is increasingly evolved, which forms a basis of common sense for humans.

## Final remarks

In this paper, we propose a design of HLR machines in the context of object identification, which is much inspired by Turing Test proposed 70 years ago. When Turing aimed at the evaluating the machines in terms of intelligence, we advance such idea on enabling machines in terms of human-like robustness. We wish the design will provide a technical solution for reaching the two goals of AI machines. We summarize two detailed contributions as well as two limitations below about the design.

Contribution 1: We define the three classes of robustness for distinguishing them in both linguistic and mathematic senses. The novel definitions will help us in building HLR AI machines, which is significantly different with the existing solutions from the low-level robustness in most studies.

Contribution 2: The design of HLR AI machines advances the idea of Turing Test by including both enabling and evaluating intelligence over machines in terms of HLR. We propose the three features as the design guidelines for realizing HLR AI machines so that the idea of HAI is advanced further.

Limitation 1: The design is very primary and staying at the high-level aspects of the models. Neither methods nor algorithms are given about the implementations of the models. We suggest readers refer to the related papers, such as the methods in common sense representations[13,14,22,23], semantic similarity (or distance)[24–26], and assessment of robustness[27].

Limitation 2: The intelligence in the design is limited only within a concept of robustness. The example is given only on a visual robustness. The human robustness covers a large spectrum in applications, such as, perception, control, decision making, etc.

More issues exists about HLR AI machines, such as "*to originate anything*"[1] for such kind of machines. Our position is to learn from Turing by proposing a novel idea first, and putting implementations or related issues in future studies. We hope that the proposal of the design is able to bring a new study space for us to explore.

In a final remark, when we express the deep respect to Alan Turing for establishing a link of humans and machines in terms of intelligence, we stress the term "**HUMANITY**" in designs and applications of human-like AI machines at the same time, and express the great respect to Confucius (孔子) (551-479 BC) in proposing humanity goal (仁者愛人, *ren* or humaneness)[28,29] for humans before two thousand years ago which we can never ignore in our future development.

(Dated: December 16, 2020)